\documentclass[letterpaper,journal]{IEEEtran}
\usepackage{amsmath,amsfonts}
\usepackage{algorithmic}
\usepackage{algorithm}
\usepackage{array}
\usepackage[caption=false,font=normalsize,labelfont=sf,textfont=sf]{subfig}
\usepackage{textcomp}
\usepackage{stfloats}
\usepackage{url}
\usepackage{verbatim}
\usepackage{multirow}
\usepackage{booktabs}
\usepackage{pifont}
\usepackage{graphicx}

\usepackage{cite}
\usepackage{etoolbox}

\newcommand{\chg}[1]{{\scriptsize\textcolor{gray}{(#1)}}}

\newcommand{\mycites}[1]{%
  \def\temp{}%
  \renewcommand{\do}[1]{%
    \ifx\temp\empty
      \cite{##1}\def\temp{nonempty}%
    \else
      , \cite{##1}%
    \fi
  }%
  \docsvlist{#1}%
}

\usepackage[colorlinks=true,      
            linkcolor=blue,      
            citecolor=blue,       
            urlcolor=blue,        
            anchorcolor=blue,
            filecolor=blue,       
            ]{hyperref}

\usepackage[table]{xcolor}
\usepackage[capitalize]{cleveref}
\crefname{section}{Sec.}{Secs.}
\crefname{table}{Table}{Tabs.}
\crefname{figure}{Fig.}{Figs.}

\begin{document}

\title{RDANet: Relative Degradation Aware Network for \\ Infrared Small Target Detection}

\author{Rui Liu, Jing Nie, and Ying Fu,~\IEEEmembership{Senior~Member,~IEEE}
\thanks{This work was supported by the National Natural Science Foundation of China under Grant 62331006 and by the Fundamental Research Funds for the Central Universities. (Corresponding author: Ying Fu.)}
\thanks{Rui Liu and Ying Fu are with the School of Computer Science and Technology, Beijing Institute of Technology, Beijing 100081, China (e-mail: liurui25@bit.edu.cn; fuying@bit.edu.cn).}%
\thanks{Jing Nie is with the School of Information and Electronics, Beijing Institute of Technology, Beijing 100081, China (e-mail: 3420235028@bit.edu.cn).}%
}

\markboth{IEEE TRANSACTIONS ON GEOSCIENCE AND REMOTE SENSING}%
{Liu \MakeLowercase{\textit{et al.}}: RDANet: Relative Degradation Aware Network for Infrared Small Target Detection}


\maketitle

\begin{abstract}

Infrared small target detection is still challenging in remote sensing imagery, because the targets are extremely small, exhibit weak local contrast, and are often embedded in complex and highly variable backgrounds. In addition to these inherent difficulties, we observe that existing detectors often show unstable performance when the target scale changes or when the scene background varies. This scale- and scene-sensitive degradation indicates that current methods are insufficient in simultaneously preserving target structure during feature downsampling and maintaining discriminative local contrast under background shifts, which finally results in unbalanced detection performance across different conditions.
To improve detection robustness, this paper proposes a Relative Degradation Aware Network (RDANet) for infrared small target detection. RDANet consists of two dedicated modules: Multi-Scale Anti-Alias Downsampling (MSAD) and Prototype-Guided Skip Memory (PGSM). MSAD introduces multi-scale anti-alias filtering together with pixel-fold aggregation to reduce aliasing effects during resolution reduction, so that target shape information can be better preserved while irrelevant background responses are suppressed. PGSM further enhances the skip features by retrieving patch-level prototypes from a shared memory and adaptively integrating them into the current representation, which helps maintain stable local contrast cues under diverse scene backgrounds.
Experiments on three public benchmarks show that RDANet achieves the best performance on most evaluation metrics, while scale- and background-stratified evaluations indicate more stable behavior across target sizes and scene complexity. The code is available at \href{https://github.com/BIT-RuiLiu/RDANet}{https://github.com/BIT-RuiLiu/RDANet}.
\end{abstract}


\section{Introduction}
\label{sec:intro}

\begin{figure}
    \centering
    \includegraphics[width=\linewidth]{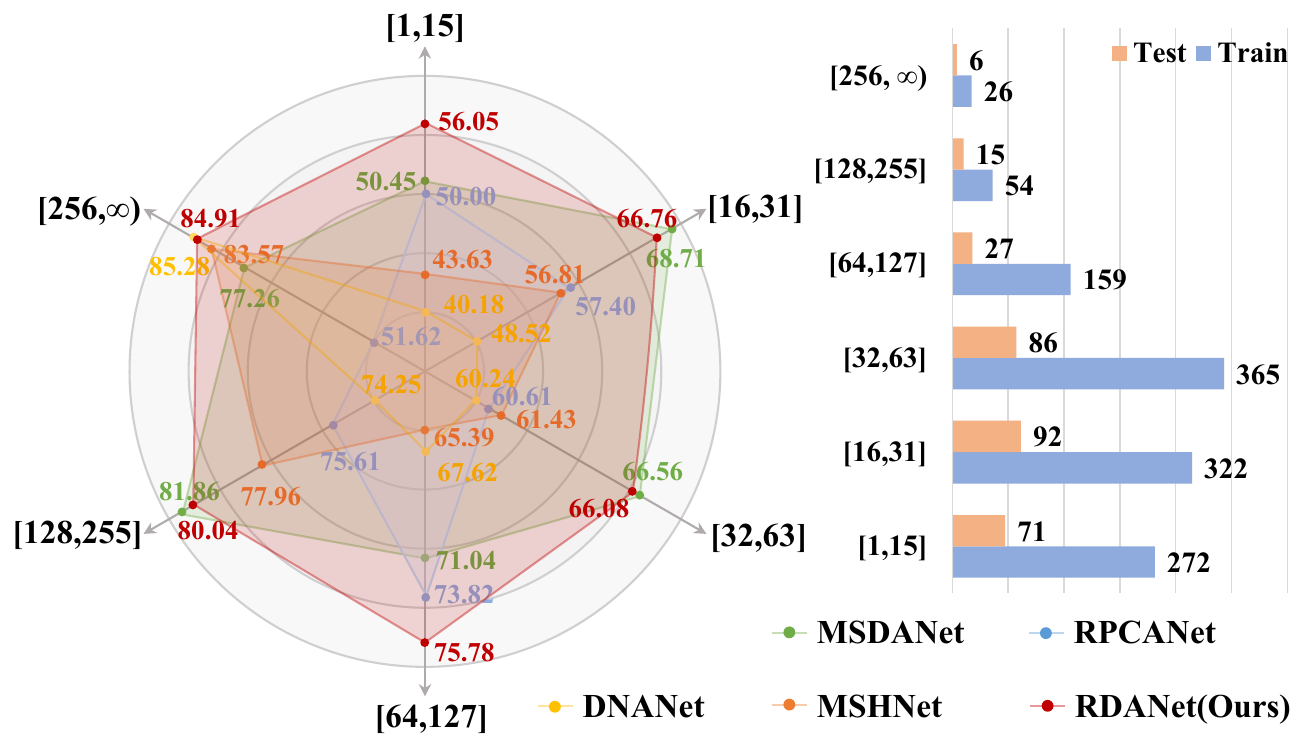}
    \caption{Scale-wise distribution and detection performance across target sizes on IRSTD-1k. The right panel shows the distribution of training and test samples across target size intervals. The left panel presents the IoU of representative IRSTD methods across six target size ranges. Most methods exhibit unstable or degraded performance as target size increases. In contrast, our RDANet maintains stable performance across scales.}
    \label{fig:teaser}
\end{figure}

\begin{figure}[t]
    \centering
    \includegraphics[width=\linewidth]{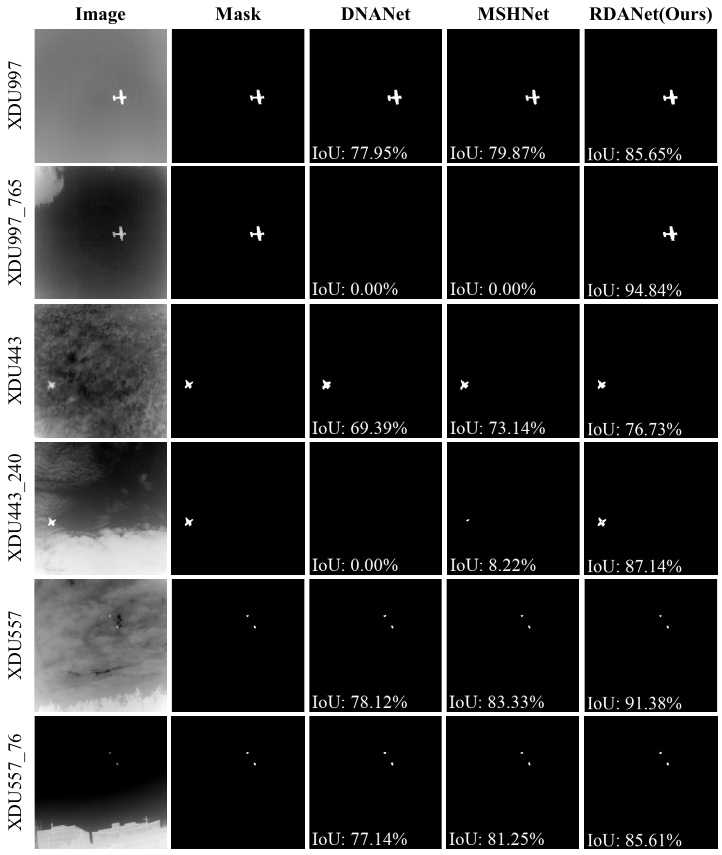}
    \caption{Detection results of representative deep-learning-based methods for targets at different scales and backgrounds. Changing the surrounding background causes pronounced prediction variations for enlarged targets, whereas the responses become more stable as the target size decreases.}
    \label{fig:degradation}
\end{figure}    

\IEEEPARstart{I}{nfrared} small target detection (IRSTD) plays an important role in a wide range of remote sensing and defense-related applications, such as aerial early warning \cite{warning_1, warning_2}, maritime surveillance \cite{maritime1, maritime_1}, and disaster rescue \cite{rescue_1, rescue_2}. In these scenarios, targets usually occupy only a few pixels and often appear with low signal-to-clutter ratios, making them highly susceptible to background interference and imaging noise. Early IRSTD methods mainly relied on hand-crafted features and manually designed priors to characterize local contrast, sparse structures, or low-rank background properties. Although such methods can achieve reasonable performance in relatively simple scenes, they often struggle when facing complex backgrounds, sensor noise, or significant variations in target appearance and scale. Benefiting from end-to-end representation learning and hierarchical feature modeling, deep-learning-based methods have substantially improved the robustness and accuracy of infrared small target detection, and have gradually become the dominant paradigm in this field. However, despite these advances, existing deep models still show clear limitations in maintaining balanced detection performance across targets of different sizes.

We observe a clear performance drop on IRSTD datasets with imbalanced target-size distributions. Specifically, when the training data is dominated by small-scale targets, existing methods often show limited generalization to relatively larger targets. As shown in the right panel of \cref{fig:teaser}, the widely used IRSTD-1k dataset \cite{zhang2022isnet} exhibits an evident scale-wise imbalance in its target distribution. To investigate how current detectors respond to this issue, we select four representative methods with different design strategies. RPCANet~\cite{wu2024rpcanet} is built on low-rank decomposition, MSDANet~\cite{zhao2025multi} adopts a multi-scale directional aggregation design, while MSHNet~\cite{MSHNet} and DNANet~\cite{dnanet} are CNN-based models with multi-scale or contrast-aware designs. As shown in the left panel of \cref{fig:teaser}, although these methods achieve competitive overall performance, their detection accuracy varies noticeably across target-size ranges. In particular, RPCANet and MSDANet generalize poorly to scale-expanded targets when such samples are insufficient in the training set, indicating that existing methods still lack robust scale-consistent representation learning.

Although MSHNet and DNANet partially alleviate this size imbalance, we further find that their performance remains sensitive to background variations. As illustrated in \cref{fig:degradation}, under a controlled setting where the target patch is kept fixed and only the surrounding scene is changed, the detection responses of scale-expanded targets fluctuate noticeably across different backgrounds. By contrast, when the target size becomes smaller, the fluctuation is gradually reduced and the responses become relatively more stable across scenes. This phenomenon suggests that the degradation of existing methods is not only related to target scale, but is also closely coupled with background-induced feature variation. Such relative degradation across scales and backgrounds reveals two key limitations of current deep IRSTD models: first, deep target semantics are not sufficiently preserved during resolution changes; second, the learned feature responses remain fragile to local contrast shifts caused by scene variation.

In this paper, we propose the Relative Degradation Aware Network (RDANet) to address the above limitations. RDANet adopts a simple encoder-decoder architecture and introduces two lightweight yet effective components, namely Multi-Scale Anti-Alias Downsampling (MSAD) and Prototype-Guided Skip Memory (PGSM). The design of RDANet is motivated by the need to improve both scale-consistent structure preservation and scene-robust local contrast modeling. Specifically, MSAD performs resolution transformation through a multi-kernel low-pass filtering path together with a pixel-fold aggregation branch. The low-pass path suppresses aliasing effects and reduces background leakage during sampling, while the pixel-fold branch reorganizes fine-grained structures to better preserve target shape across scales. In parallel, PGSM enhances skip features by retrieving patch-wise prototypes from a shared memory bank and adaptively fusing them with learnable weights. This prototype-guided fusion helps maintain stable local contrast under scene changes and strengthens target-related responses without introducing excessive background interference. With these two components, RDANet is able to regularize feature representation across target scales and maintain more consistent detection responses under varying backgrounds. Experiments on three public benchmarks show that RDANet achieves competitive overall performance and more stable results across target scales and background conditions.

Our main contributions are summarized as follows:
\begin{itemize}
\item We analyze a relative degradation phenomenon in IRSTD, where the performance of existing detectors becomes unstable across target scales and background changes, and accordingly propose RDANet to improve robustness under such conditions.
\item We design MSAD, which combines multi-kernel anti-alias filtering with pixel-fold aggregation to preserve target structure and suppress background leakage during resolution transformation.
\item We introduce PGSM, which retrieves patch-wise prototypes from a shared memory and adaptively fuses them into skip features to maintain more stable local contrast cues across scenes.
\end{itemize}

\section{Related Work}
\label{sec:related}

\subsection{Infrared Small Target Detection}
Existing IRSTD methods can be divided into traditional and deep learning approaches. Traditional methods include filtering-based \cite{tophat, max_mean}, local-contrast \cite{wslcm, tllcm}, and low-rank decomposition methods \cite{ipi, PSTNN, MSLSTIPT}. Although lightweight and interpretable, they rely heavily on hand-crafted priors and often degrade under low contrast, heavy clutter, and scene variation. Deep learning methods have become dominant due to their end-to-end and hierarchical representation capabilities \cite{alcnet, Li2023FCDFusion, zhang2022isnet, zhang2026supervise, zhang2025unaligned, dnanet, zhang2022rkformer, TriMSOD, CJE1, CJE2, chen2026survey, fang2023dual, zhang2024deep, jiang2026msfa}. They have evolved from shallow networks \cite{liuinfrared} to GAN-based \cite{mdvsfa}, nested U-shaped \cite{uiunet}, SAM-style \cite{zhang2025irsam}, Mamba-based \cite{chen2024mim, zhang2025irmamba}, and text-guided models \cite{zhang2025saist, huang2025text}. However, improved accuracy is often accompanied by higher complexity and limited robustness to scale and scene variations.

Recent studies have therefore focused on scale-aware modeling and robustness enhancement. For scale variation, Liu \textit{et al.}~\cite{liu2024mshnet} combine an SLS loss with a multi-scale head, while Dai \textit{et al.}~\cite{dai2024serankdet} employ selective rank-aware attention and dynamic feature fusion. Yang \textit{et al.}~\cite{yang2025pconv} integrate pinwheel-shaped convolution with a scale-based dynamic loss. Other methods improve scale-aware representations through dynamic local context modeling \cite{zhang2025lcrnet}, multi-scale context aggregation \cite{lu2025mscanet}, and scale-aligned fusion \cite{wang2025mafnet}.

For scene robustness, Zhao \textit{et al.}~\cite{zhao2025fest} enhance multi-scale features and regulate confidence to improve adaptability in complex scenes. Lu \textit{et al.}~\cite{lu2025realscene} introduce a real-scene benchmark and cross-view alignment for cross-scene generalization, while Xiong \textit{et al.}~\cite{xiong2025drpcanet} generate scene-conditioned parameters through dynamic RPCA unfolding. Zhang \textit{et al.}~\cite{zhang2025lrrnet} improve resilience to noise and clutter through deep patch-free low-rank modeling. Zhang \textit{et al.}~\cite{zhang2024irprunedet} improve model efficiency through wavelet-structure-regularized pruning. Feature compensation and cross-level correlation \cite{zhang2022exploring} further strengthen multilevel representations. However, robustness to scale and background variations remains insufficiently explored. Consequently, existing methods still have difficulty achieving stable performance across both target scales and background conditions.

\subsection{Downsampling for IRSTD}
Downsampling expands the receptive field and enables hierarchical representation learning, but conventional pooling and strided convolutions may introduce aliasing and background leakage. These effects can easily suppress weak target cues, particularly for tiny targets or strong clutter. To alleviate this problem, some recent studies have explored alternative downsampling designs. Building on pixel rearrangement strategies \cite{shuffle}, Sunkara and Luo \cite{sunkara2022no} propose SPDConv, which preserves fine-grained spatial details by rearranging local information into the channel dimension before convolution. This idea has been introduced into IRSTD by subsequent works \cite{liu2023glcanet, qi2025investigation}, which report improved detection accuracy at the cost of noticeably increased computation. Fan \textit{et al.} \cite{fan2024improved} adopt Haar wavelet downsampling \cite{xu2023haar} to reduce feature loss and better preserve target structures, while Xu \textit{et al.} \cite{xu2024single} insert BlurPool before stride operations to mitigate positional bias and improve localization stability. Although these variants improve feature preservation to some extent, they mainly focus on reducing information loss during sampling and do not explicitly address the preservation of large-target morphology under complex backgrounds. As a result, their robustness under imbalanced target-size distributions remains limited.

\subsection{Memory-Based Enhancement for IRSTD}
Memory mechanisms enhance IRSTD by storing and retrieving informative patterns for feature recovery, temporal consistency, and prototype learning. RPCANet++ \cite{wu2025rpcanet++} introduces a memory-augmented block to recover target-related features under complex backgrounds, thereby improving target-background separation after RPCA unfolding. Duan \textit{et al.} \cite{duan2024triple} develop a frequency-aware memory enhancement strategy within a triple-domain framework for infrared small target detection in video sequences, which strengthens cross-frame consistency for dim targets. Prototype-based methods \cite{duan2025semi} further exploit multi-view information to enhance robustness against limited annotations and motion variation in moving-target scenarios. Related temporal methods, including temporally aware FCNs and motion-perception based frameworks \cite{zhang2023infrared}, also demonstrate the value of retaining and recalling historical context for suppressing false alarms and stabilizing target responses across frames. Nevertheless, existing methods mainly emphasize temporal persistence, global recovery, or sequence-level consistency. Patch-wise prototype retrieval within skip connections and its coordination with downsampling remain largely unexplored, leaving scale imbalance and background-induced contrast shifts insufficiently addressed.

\section{Methodology}
\subsection{Overview of RDANet}
Although recent IRSTD methods have achieved notable progress, their performance is still limited by insufficient stability across target scales and scene conditions. As discussed in the Introduction, existing detectors often degrade when the target scale expands beyond the dominant training distribution, and their responses may further fluctuate under background changes. Rather than treating this phenomenon as an isolated failure of specific architectures, we consider it a more general limitation in current deep IRSTD pipelines: the feature hierarchy is not sufficiently robust to preserve target structure during resolution transformation, nor is it reliable enough to maintain stable local contrast under scene variation.

From a modeling perspective, these two issues are closely related to the encoder--decoder process. In the encoder, repeated downsampling enlarges the receptive field and aggregates contextual information, but it may also introduce aliasing, blur target morphology, and leak background interference into deeper features. This problem becomes more evident when target scales vary, since the structural integrity of relatively larger targets can be more easily disturbed by inappropriate resolution reduction. In the decoder, skip connections are expected to provide fine-grained spatial details for target recovery, yet their effectiveness is still sensitive to scene-dependent contrast variation. When local background statistics change, shallow features may drift together with the surrounding clutter, making target cues unstable even if the target itself remains unchanged. Therefore, improving scale robustness and scene robustness requires not only stronger representation learning in general, but also more dedicated designs at these two critical stages.

\begin{figure*}[ht]
    \centering
    \includegraphics[width=\linewidth]{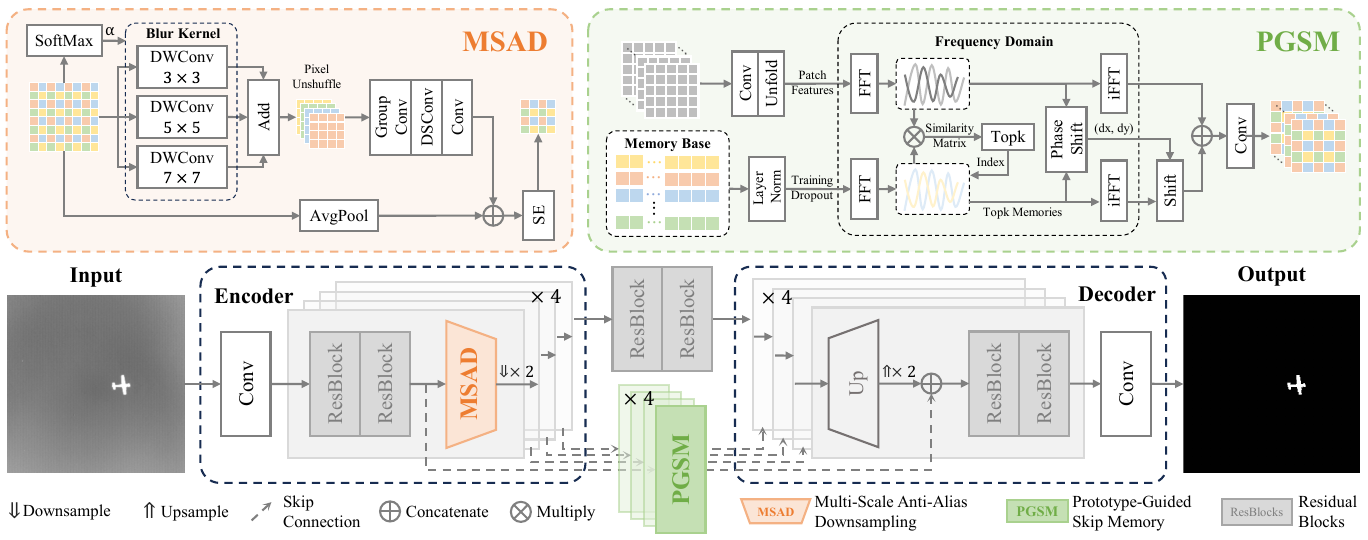}
    \caption{Overall architecture of the proposed RDANet. MSAD performs anti-aliased downsampling to preserve target structure and suppress background leakage, while PGSM enhances skip connections to stabilize local contrast under scene variations.}
    \label{fig:overview}
\end{figure*}

Based on this observation, we propose RDANet, a simple encoder--decoder framework equipped with two lightweight components to enhance feature stability throughout the network. As illustrated in \cref{fig:overview}, RDANet replaces each stride-two downsampling operation in the encoder with MSAD, which aims to preserve target morphology while suppressing aliasing and background leakage during resolution changes. In addition, each skip connection is refined by PGSM before being fused with the upsampled decoder features. PGSM enhances scene-invariant local contrast by retrieving representative patch-wise prototypes from a shared memory and adaptively integrating them into the skip pathway. Through these two complementary designs, RDANet improves both structural consistency across scales and response stability across scenes, thereby alleviating the relative degradation observed in existing IRSTD methods.

\subsection{Multi-Scale Anti-Alias Downsampling}
Although downsampling enlarges the receptive field and builds hierarchical features, conventional strided convolutions and pooling may introduce aliasing, blur target boundaries, distort target morphology, and amplify background responses. This is particularly harmful in IRSTD, where local target cues are weak. MSAD preserves target shape during resolution transformation while suppressing background texture leakage.

Given an input feature map $\mathbf{X}\in\mathbb{R}^{C\times H\times W}$, MSAD first performs multi-scale low-pass filtering before spatial reduction. Specifically, we apply three fixed binomial kernels $\{K_3, K_5, K_7\}$ with kernel sizes $3$, $5$, and $7$ as depthwise convolutions to obtain three scale-specific responses $\{\mathbf{Y}_3, \mathbf{Y}_5, \mathbf{Y}_7\}$. These filters provide different smoothing ranges and allow the module to adaptively handle target and background structures of different spatial extents. A channel-wise attention gate is then used to predict the fusion weights
$\boldsymbol{\alpha}\in\mathbb{R}^{3\times C\times 1\times 1}$ through a softmax activation, and the filtered responses are aggregated as
\begin{equation}
\mathbf{Y}=\alpha_3\odot \mathbf{Y}_3+\alpha_5\odot \mathbf{Y}_5+\alpha_7\odot \mathbf{Y}_7,
\end{equation}
where $\odot$ denotes element-wise multiplication. In this way, each channel can adaptively select an appropriate smoothing scale according to its content, which helps suppress high-frequency noise and background interference before downsampling.
The three low-pass ranges play complementary roles: smaller kernels preserve weak target details, whereas larger kernels suppress a broader range of high-frequency clutter. Adaptive channel-wise fusion balances these responses instead of applying one fixed smoothing strength to all features.

After anti-alias filtering, we apply a pixel-unshuffle operator with stride $2$, denoted by $\mathrm{PixelUnshuffle}(\cdot)$, to reorganize each local $(2\times2)$ neighborhood into the channel dimension:
\begin{equation}
\mathbf{Y}'=\mathrm{PixelUnshuffle}(\mathbf{Y})\in\mathbb{R}^{4C\times \frac{H}{2}\times \frac{W}{2}}.
\end{equation}
Unlike direct strided sampling, this operation retains local sub-pixel information during resolution reduction and avoids discarding fine-grained target structures. To restore the channel dimension, a grouped convolution with fixed weights ($1/4$ for each group) is further applied to average every four folded sub-channels:
\begin{equation}
\mathbf{Y}''=\mathrm{GroupConv}(\mathbf{Y}')\in\mathbb{R}^{C\times \frac{H}{2}\times \frac{W}{2}}.
\end{equation}
This process can be viewed as a parameter-free sub-pixel aggregation strategy, which preserves local structural continuity without introducing unnecessary cross-channel mixing.

To further enhance the transformed features, we employ a depthwise separable convolution consisting of a depthwise $3\times3$ convolution followed by a pointwise $1\times1$ convolution:
\begin{equation}
\tilde{\mathbf{Y}}=\mathrm{Conv}_{1\times1}\big(\mathrm{Conv}_{3\times3}(\mathbf{Y}'')\big).
\end{equation}
This design introduces local feature refinement with low computational overhead, allowing the module to enhance target-related responses after downsampling while maintaining efficiency.

Meanwhile, a residual shortcut is constructed by directly downsampling the input feature map $\mathbf{X}$ through average pooling, and then combining it with the refined main branch:
\begin{equation}
\mathbf{Z}=\mathrm{ReLU}\big(\tilde{\mathbf{Y}}+\mathrm{AvgPool}_2(\mathbf{X})\big).
\end{equation}
The residual path provides a stable low-frequency reference and helps preserve the overall feature distribution during resolution change, thereby improving optimization and preventing excessive information loss.

Finally, we apply channel recalibration using an SE block \cite{se} to further emphasize target-relevant channels and suppress residual background activations:
\begin{equation}
\mathrm{out}=\mathbf{Z}\odot \sigma\left(\mathrm{Conv}_{1\times1}\big(\mathrm{ReLU}(\mathrm{Conv}_{1\times1}(\mathrm{GAP}(\mathbf{Z}))))\right),
\end{equation}
where $\mathrm{GAP}(\cdot)$ denotes global average pooling and $\sigma(\cdot)$ denotes the sigmoid function. Through this final channel-wise modulation, MSAD adaptively strengthens informative responses and improves the robustness of downsampled features.

Overall, MSAD combines multi-scale anti-alias filtering, sub-pixel structure-preserving aggregation, residual compensation, and channel recalibration within a lightweight module. By explicitly reducing aliasing and preserving target morphology during resolution transformation, it provides more stable feature representations for subsequent encoding stages and improves robustness under target-scale variation and complex background conditions.

\subsection{Prototype-Guided Skip Memory}
Skip connections recover fine spatial details, but in IRSTD, shallow features are sensitive to scene-dependent background variations even for similar targets, causing feature drift and weakening target–background separation. PGSM mitigates this by retrieving representative patch-wise patterns from a learnable memory bank and injecting scene-stable prototypes into the skip pathway, thereby preserving consistent local contrast cues across backgrounds.

Given an input skip feature map $\mathbf{F}\in\mathbb{R}^{C\times H\times W}$, we first construct a query representation for patch-wise retrieval. Specifically, $\mathbf{F}$ is projected into a reduced channel space by a $1\times1$ convolution, and then unfolded into a set of patch tokens
$\mathbf{Q}\in\mathbb{R}^{B\times N\times D}$,
where $N$ denotes the number of patches and
$D=C'P^2$ is the token dimension determined by the projected channel number $C'$ and patch size $P$. This patch-wise tokenization allows the memory module to model local target and background structures at a finer granularity, instead of relying only on global feature statistics.

To provide reusable local patterns, we maintain a global learnable memory bank
$\mathbf{M}\in\mathbb{R}^{M\times D}$,
where each memory slot stores a prototypical feature patch that summarizes representative local structures observed across scenes. Given a query token $\mathbf{q}_i\in\mathbb{R}^{D}$ and a memory token $\mathbf{m}_j\in\mathbb{R}^{D}$, PGSM computes their similarity using a frequency-aware cosine measure. Concretely, both tokens are reshaped into $C'\times P\times P$ patches, modulated by a Hann window, and transformed into the frequency domain. Their log-transformed magnitude spectra are then extracted through cropped 2D FFT:
\begin{equation}
\hat{\mathbf{q}}_i = \mathrm{FFTmag}(\mathbf{q}_i), \quad
\hat{\mathbf{m}}_j = \mathrm{FFTmag}(\mathbf{m}_j),
\end{equation}
where $\mathrm{FFTmag}(\cdot)$ denotes Hann-windowing followed by low-frequency cropping and log-magnitude extraction. Based on these spectral representations, the similarity is defined as
\begin{equation}
\mathrm{sim}(\mathbf{q}_i,\mathbf{m}_j)=
\frac{\langle \hat{\mathbf{q}}_i,\hat{\mathbf{m}}_j\rangle}
{\|\hat{\mathbf{q}}_i\|_2\cdot\|\hat{\mathbf{m}}_j\|_2},
\end{equation}
where $\langle\cdot,\cdot\rangle$ denotes the inner product and $\|\cdot\|_2$ denotes the $\ell_2$ norm. Compared with direct spatial matching, this frequency-aware formulation is less sensitive to small positional deviations and emphasizes structural magnitude patterns that are more stable across scenes.

For each query token, the top-$K$ memory slots with the highest similarity scores are selected and retrieved. Since the matched prototypes may still exhibit local spatial offsets relative to the current query patch, we further estimate a sub-pixel displacement by phase correlation and align the retrieved memory patches accordingly. The aligned memory features are then aggregated through attention weighting:
\begin{equation}
\mathbf{g}_i = \sum_{k=1}^{K}\alpha_{ik}\cdot \mathrm{Warp}(\mathbf{m}_{ik}),
\end{equation}
where $\alpha_{ik}$ are normalized attention weights and $\mathrm{Warp}(\cdot)$ denotes bilinear warping based on the estimated offset. This alignment-and-aggregation process allows PGSM to retrieve not only semantically relevant prototypes, but also spatially compatible local patterns, which is important for preserving target continuity in cluttered scenes.

Unlike memory modules for global restoration or temporal aggregation, PGSM performs patch-wise prototype retrieval directly on skip features. Frequency-magnitude matching reduces sensitivity to local shifts and background texture changes, while phase-correlation alignment compensates residual offsets before fusion. The retrieved prototype acts as a structural reference, whereas the original skip feature remains the primary source of target detail.

After retrieval, all aggregated patch features are folded back to the spatial domain to obtain a full-resolution memory-enhanced feature map $\mathbf{G}$. We then concatenate $\mathbf{G}$ with the original skip feature $\mathbf{F}$ and fuse them through a $1\times1$ convolution followed by BatchNorm and ReLU:
\begin{equation}
\mathrm{out}=\mathrm{ReLU}\big(\mathrm{BN}(\mathrm{Conv}_{1\times1}([\mathbf{F},\mathbf{G}]))\big),
\end{equation}
where $[\cdot,\cdot]$ denotes channel-wise concatenation. Through this fusion, the original skip pathway is enriched by context-aware prototype cues, enabling the network to suppress scene-specific background interference while retaining discriminative target-related details.

PGSM serves as a lightweight memory-enhanced skip refinement module. By retrieving and aligning representative patch-wise prototypes, it improves the stability of local contrast modeling under background changes and provides more reliable fine-grained features for the decoder. 

\begin{table*}[!t]
\begin{center}
\setlength{\tabcolsep}{9pt}
\small
\caption{Quantitative comparison of different methods in terms of IoU (\%), ${\rm P_d}$ (\%), and ${\rm F_a}$ ($10^{-6}$). The best and second-best results are highlighted in \textbf{bold} and \underline{underlined}, respectively.}
\resizebox{\textwidth}{!}{%
\begin{tabular}{l|l|ccc|ccc|ccc}
\toprule[1pt]
\multirow{2}{*}{Method} & \multirow{2}{*}{Venue}  & \multicolumn{3}{c|}{IRSTD-1k} & \multicolumn{3}{c|}{NUDT-SIRST} & \multicolumn{3}{c}{NUAA-SIRST}  \\ \cline{3-11}
& & IoU$\uparrow$ & ${\rm P_d}\uparrow$ & ${\rm F_a}\downarrow$ & IoU$\uparrow$ & ${\rm P_d}\uparrow$ & ${\rm F_a}\downarrow$ & IoU$\uparrow$ & ${\rm P_d}\uparrow$ &  ${\rm F_a}\downarrow$ \\ \midrule

Top-Hat\cite{tophat}  & OE'96 & 10.06 &  75.11 &  1432 & 20.72  & 78.41 & 166.70 &  1.51 & 79.74 &  16456   \\
WSLCM\cite{wslcm} & GRSL'20 &  3.45 & 72.44 & 6619 & 2.28 & 56.82  & 1309 &  6.39  & 88.74 & 4462 \\
TLLCM\cite{tllcm} & GRSL'19 & 3.31 & 77.39 & 6738 & 2.28 & 62.01  & 1608 &  4.24 & 88.37 & 6243   \\
IPI\cite{ipi} & TIP'22 & 27.92 & 81.37 & 16.18 & 17.76 & 74.49 & 41.23 & 1.09 & 87.05 & 30467  \\
NRAM\cite{nram} & RS'18 & 15.25 & 70.68 & 16.93 & 6.93 & 56.40 & 19.27 & 13.54 & 60.04 & 25.23  \\
RIPT\cite{RIPT} & JSTARS'17 &  14.11 & 77.55 & 28.31 & 29.44 & 91.85 & 344.30 & 16.79 & 69.76 & 59.33  \\
PSTNN\cite{PSTNN} & RS'19 &  24.57 & 71.99 & 35.26 & 14.85 & 66.13 & 44.17 & 30.30 & 72.80 & 48.99  \\
\midrule
ACM\cite{acmnet} & WACV'21 & 60.21 & 84.18 & 24.82 & 72.33 & 89.58 & 17.85 & 68.31 & 91.63 & 18.43 \\
DNANet\cite{dnanet} & TIP'22 & 65.71 & 91.84 & 17.61  & 94.19 & \underline{99.26} & \underline{2.44} & 74.82 & 93.54 & 38.28 \\
ISNet\cite{zhang2022isnet} & CVPR'22  & 61.85 & 90.24 & 31.56 & 81.24 & 97.78 & 6.34 & 70.49 & 95.06 & 67.98  \\
UIUNet\cite{uiunet} & TIP'23 & 65.69 & 91.25 & 13.48 & 90.52 & 98.84 & 8.34 & 77.53 & 92.40 & 9.33  \\ 
MSHNet\cite{MSHNet} & CVPR'24 & 67.87  & 92.86  & 8.88 & 80.55  & 97.99 & 11.77 & 72.85  & 97.25 &  28.57  \\
SCTransNet\cite{yuan2024sctransnet} & TGRS'24 & 68.03 & 93.27 & 10.74 & 94.09 & 98.62 & 4.29 & 77.50 & 96.95 & 13.92 \\
RPCANet\cite{wu2024rpcanet} & WACV'24 & 63.50 & 87.63 & \underline{8.65} & 89.37  & 95.76 & 18.57 & 70.41  & 94.50 & 8.69  \\
L$^2$SKNet\cite{wu2025lk} & TGRS'25 & 67.81 & 90.24 & 17.46 & 93.58 & 97.57 & 5.33 & 73.43 & 98.17 & 20.82  \\
MMLNet\cite{li2025multi} & TGRS'25 & 67.21 & \textbf{94.28} & 14.00  & 81.81 & 98.43 & 11.77 & 78.71 & \underline{98.43} & 11.77   \\
MSDANet\cite{zhao2025multi}  & TGRS'25 & \underline{71.89} & \textbf{94.28} & 11.39  & 93.81 & 99.19 & 3.70 & \textbf{81.08} & \textbf{100.00} & 7.19  \\
Text-IRSTD \cite{huang2025text} & ICCV'25 &  65.50 & 91.56  & 16.13 &  \underline{95.25} & 98.94 & \textbf{1.66} & -- & -- & -- \\
\rowcolor{gray!20}
RDANet & Ours   & \textbf{73.82}  & \underline{93.60} & \textbf{7.67} & \textbf{95.43} & \textbf{99.47} & 2.90 &  \underline{79.33} & \textbf{100.00} & \textbf{1.91} \\
 \bottomrule[1pt]                 
\end{tabular}
}
\label{tab:main_result}
\end{center}
\end{table*}

\begin{table}[!t]
\centering
\small
\caption{Scale-wise IoU(\%) comparison on NUDT-SIRST.}
\setlength{\tabcolsep}{5pt}
\resizebox{\columnwidth}{!}{%
\begin{tabular}{l|cccc|c}
\toprule[1pt]
Method & [1,15] & [16,31] & [32,63] & [64,127] & IoU$\uparrow$ \\
\midrule
MSHNet\cite{MSHNet}              
& 87.59 & 67.87 & 84.34 & 83.71 & 80.55 \\
DNANet\cite{dnanet}              
& 92.46 & \underline{93.13} & \underline{94.97} & 94.41 & \underline{94.19} \\
RPCANet\cite{wu2024rpcanet}      
& 89.15 & 85.27 & 89.69 & 91.95 & 89.37 \\
MSDANet\cite{zhao2025multi} 
& \textbf{96.34} & 91.74 & 93.72 & \underline{94.57} & 93.81 \\
\rowcolor{gray!20}
RDANet (Ours)                    
& \underline{94.13} & \textbf{93.88} & \textbf{95.94} & \textbf{96.34} & \textbf{95.43} \\
\bottomrule[1pt]
\end{tabular}
}
\label{tab:scale_result}
\end{table}

\section{Experiments}
\subsection{Experimental Details}
\paragraph{Datasets}
We evaluate the proposed RDANet on three widely used public IRSTD benchmarks, namely IRSTD-1k \cite{zhang2022isnet}, NUDT-SIRST \cite{dnanet}, and NUAA-SIRST \cite{acmnet}. Specifically, IRSTD-1k contains 1,001 infrared images, NUDT-SIRST contains 1,327 images, and NUAA-SIRST includes 427 images. To ensure a fair comparison and consistent training setting across datasets, all images are resized to $512\times512$. Following the standard experimental protocols used in prior works \cite{zhang2022isnet, dnanet, acmnet}, IRSTD-1k and NUAA-SIRST are divided into training and testing sets with a ratio of 4:1, while NUDT-SIRST adopts a 1:1 split.

\paragraph{Evaluation Metrics}
We adopt both pixel-level and object-level metrics to comprehensively evaluate detection performance. At the pixel level, we use intersection over union ($\rm IoU$) to measure the overlap between the predicted mask and the ground-truth target region. At the object level, we report detection probability ($\rm P_d$) and false alarm rate ($\rm F_a$), which reflect the ability of a method to correctly detect targets while suppressing false responses. In addition, to assess model efficiency and computational complexity, we report the number of parameters (Params) and floating-point operations (FLOPs).

For scale robustness, IRSTD-1k targets are grouped by mask area into $[1,15]$, $[16,31]$, $[32,63]$, $[64,127]$, $[128,255]$, and $[256,\infty)$ pixels. We report $\mathrm{SB\text{-}IoU}=\frac{1}{6}\sum_{i=1}^{6}\mathrm{IoU}_i$ and the average method rank over the six bins. For background robustness, local complexity is computed in an annular neighborhood $\Omega_n$ as $C_n=\operatorname{Var}(\mathbf{I}_{\Omega_n})+\lambda\operatorname{Mean}(|\nabla \mathbf{I}_{\Omega_n}|)$. Samples are divided at the 33.33rd and 66.67th percentiles into low-, medium-, and high-complexity groups, and $\mathrm{BG\text{-}Gap}=\max_g\mathrm{IoU}_g-\min_g\mathrm{IoU}_g$.

\paragraph{Implementation Details}
RDANet is implemented in PyTorch and trained on a single NVIDIA GeForce RTX 3090 GPU. We train the network for 1,000 epochs using the AdamW optimizer together with a cosine annealing learning rate schedule. The initial learning rate is set to $1\times10^{-3}$, and the batch size is set to 4. 

\begin{table}[!t]
\centering
\small
\setlength{\tabcolsep}{2pt}
\caption{Ablation study on RDANet. The top block reports the overall contribution of MSAD and PGSM, while the bottom blocks show fine-grained ablations of the two modules.}
\label{tab:ablation_msad_pgsm}
\resizebox{\columnwidth}{!}{%
\begin{tabular}{l|l|ccccc}
\toprule[1pt]
Group & Variant & IoU$\uparrow$ & ${\rm P_d}\uparrow$ & ${\rm F_a}\downarrow$ & Params & FLOPs \\
\midrule
\multirow{4}{*}{Overall}
& \cellcolor{gray!20}Full RDANet
& \cellcolor{gray!20}\textbf{73.82}
& \cellcolor{gray!20}\textbf{93.60}
& \cellcolor{gray!20}\textbf{7.67}
& \cellcolor{gray!20}4.17
& \cellcolor{gray!20}25.24 \\
& w/o PGSM                    & 71.32 & 92.86 & 11.26 & 4.13 & 25.12 \\
& w/o MSAD                    & 69.32 & 92.59 & 12.56 & 4.10 & 24.58 \\
& w/o MSAD, PGSM              & 66.93 & 91.56 & 20.14 & 4.07 & 24.46 \\
\midrule
\multirow{2}{*}{MSAD}
& w/o PixelUnshuffle          & 72.40 & 92.59 & 11.17 & 4.17 & 25.23 \\
& w/o SE gate                 & 73.10 & 92.93 & 10.74 & 4.16 & 25.24 \\
\midrule
\multirow{2}{*}{PGSM}
& w/o shift $(dx,dy)$         & 72.60 & 92.93 & 11.86 & 4.17 & 25.24 \\
& w/o memory dropout          & 72.35 & 93.27 & 13.02 & 4.17 & 25.24 \\
\bottomrule[1pt]
\end{tabular}
}
\end{table}

\begin{figure*}[ht]
    \centering
    \includegraphics[width=\linewidth]{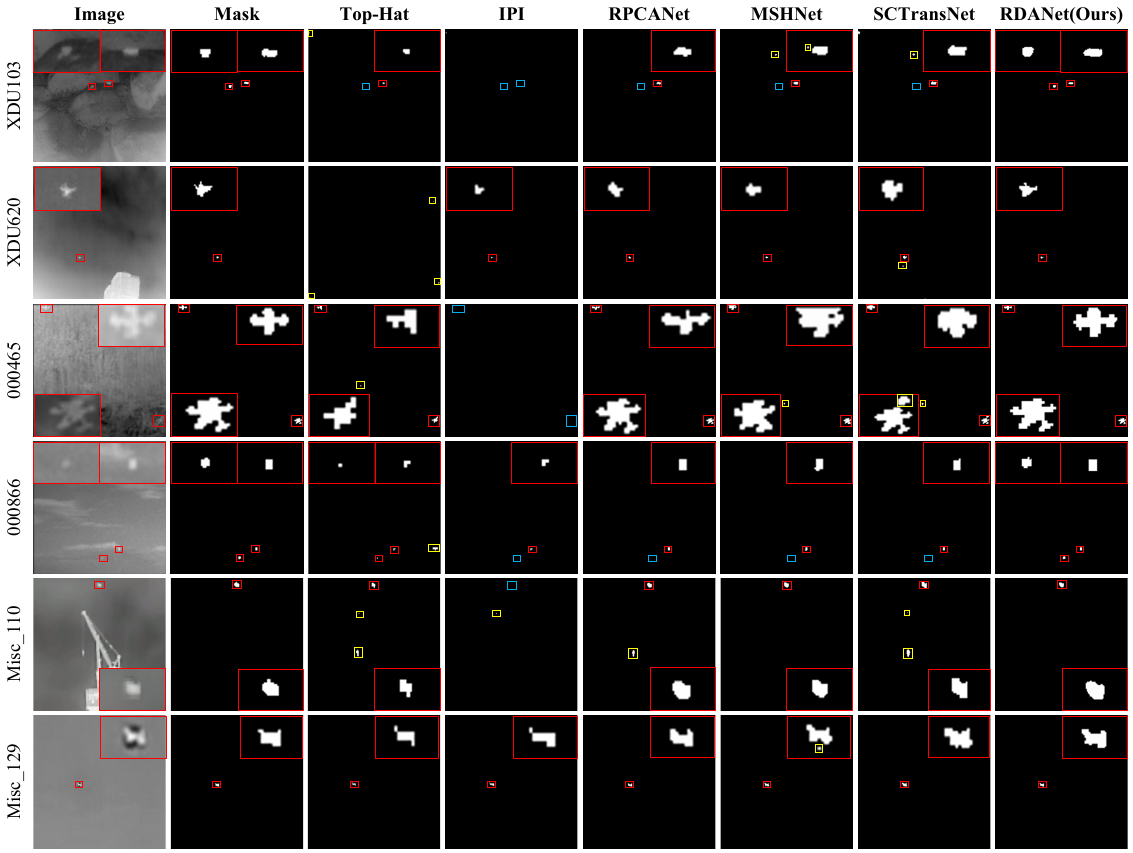}
    \caption{Visual comparisons of multiple methods are presented. Targets that are correctly detected are highlighted with red boxes, missed targets are marked with blue boxes, and false alarms are enclosed in yellow boxes. Furthermore, close-up views of the targets are shown in the corners of the images.}
    \label{fig:visual_results}
\end{figure*}

\subsection{Comparison with the State-of-the-Art Methods}
We compare the proposed RDANet with a diverse set of representative state-of-the-art IRSTD methods, including both traditional model-driven approaches and recent deep-learning-based approaches. For traditional methods, we consider Top-Hat \cite{tophat}, WSLCM \cite{wslcm}, TLLCM \cite{tllcm}, IPI \cite{ipi}, NRAM \cite{nram}, RIPT \cite{RIPT}, and PSTNN \cite{PSTNN}. For deep-learning-based methods, we include ACM \cite{acmnet}, DNANet \cite{dnanet}, ISNet \cite{zhang2022isnet}, UIUNet \cite{uiunet}, MSHNet \cite{MSHNet}, SCTransNet \cite{yuan2024sctransnet}, RPCANet \cite{wu2024rpcanet}, L$^2$SKNet \cite{wu2025lk}, MMLNet \cite{li2025multi}, MSDANet \cite{zhao2025multi}, and Text-IRSTD \cite{huang2025text}.

\paragraph{Quantitative Results}
The overall comparison is reported in \cref{tab:main_result}. RDANet achieves the highest IoU on IRSTD-1k (73.82\%) and NUDT-SIRST (95.43\%). On NUAA-SIRST, although its IoU is slightly lower than that of MSDANet, RDANet reaches ${\rm P_d}=100\%$ and the lowest ${\rm F_a}$ of 1.91. These results demonstrate a favorable balance among segmentation accuracy, detection probability, and false-alarm suppression.

Scale-stratified results are presented in \cref{tab:scale_result,tab:scale_irstd1k}. On NUDT-SIRST, RDANet ranks first in three of the four scale intervals and achieves the highest overall IoU of 95.43\%. On IRSTD-1k, it obtains the highest SB-IoU of 71.60\% and the best average rank of 1.67, indicating that its improvement is not limited to a particular target-size range.

Background-stratified results on IRSTD-1k are reported in \cref{tab:bg_irstd1k}. RDANet achieves the highest mean IoU of 73.82\% and the smallest BG-Gap of 3.97, while ranking first in both the low- and high-complexity groups. These results indicate more stable performance under varying background complexity.

\paragraph{Visual Results}
We qualitatively compare RDANet with representative traditional and deep-learning-based methods on challenging cases in \cref{fig:visual_results}. 
Traditional approaches are highly sensitive to background noise and clutter, and thus often produce fragmented responses or fail to localize targets reliably. 
deep-learning-based methods generally perform better, yet they still face two common issues in complex infrared scenes: (i) missing extremely small or low-contrast targets, and (ii) producing false positives when salient background structures resemble target-like patterns.

In contrast, RDANet yields more accurate and complete predictions across diverse scenarios. 
For example, in the first row, only RDANet successfully detects both small targets on the textured stone surface, while other methods miss the smaller one, indicating improved sensitivity to weak target cues under strong texture interference. 
In the sixth row, RDANet produces the most accurate and complete target shape among all compared methods, suggesting that it better preserves fine structural details instead of over-smoothing or fragmenting the target region. 
Moreover, as shown in \cref{fig:degradation}, RDANet maintains stable responses for the same target when composited onto different background scenes, whereas several baselines exhibit noticeable fluctuations. 
This qualitative evidence is consistent with the quantitative findings and further demonstrates RDANet's robustness to background-induced contrast shifts. Overall, the visual comparisons highlight RDANet's ability to suppress background interference while preserving target morphology in complex infrared scenes.

\begin{figure}[ht]
    \centering
    \includegraphics[width=\linewidth]{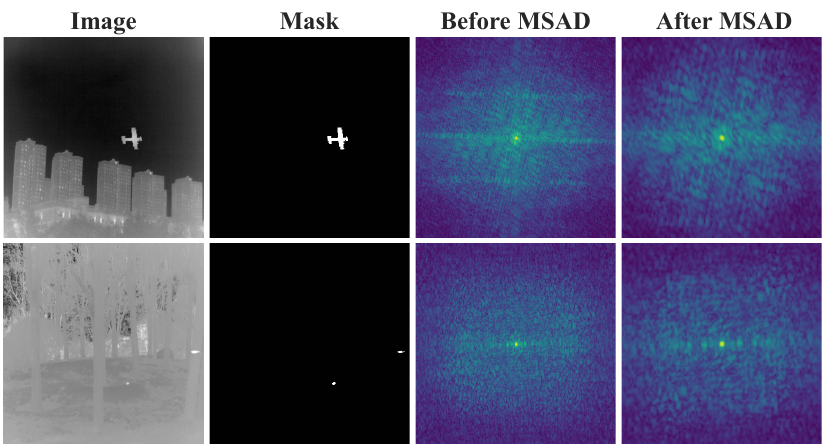}
    \caption{Frequency-spectrum visualization before and after MSAD. MSAD suppresses peripheral high-frequency components while preserving the dominant low-frequency structure.}
    \label{fig:msad_vis}
\end{figure}

\begin{figure}[ht]
    \centering
    \includegraphics[width=\linewidth]{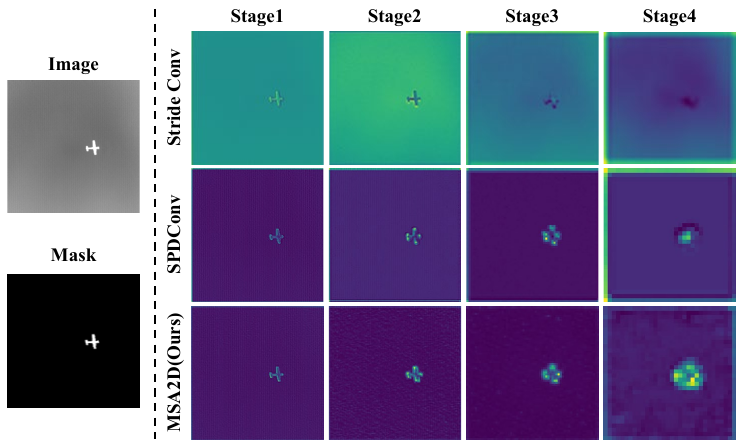}
    \caption{Feature visualizations of different downsampling methods across encoder stages. MSAD preserves clearer and more compact target responses, effectively maintaining structural integrity.}
    \label{fig:down_feature}
\end{figure}

\begin{figure}[ht]
    \centering
    \includegraphics[width=\linewidth]{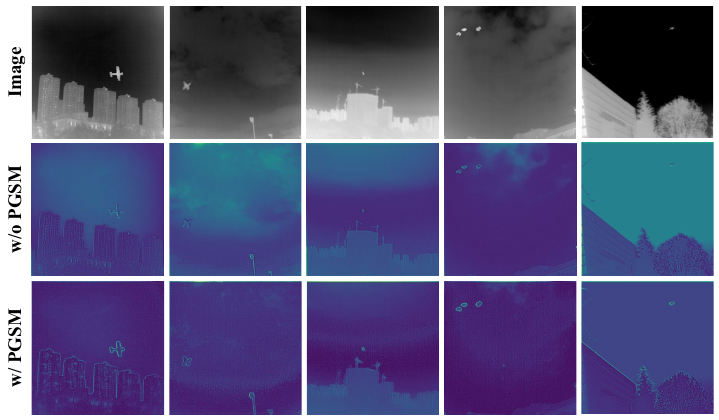}
    \caption{Visualization of feature maps before and after PGSM enhancement. PGSM significantly amplifies target activations and highlights clearer target boundaries, providing more distinctive and shape-consistent representations.}
    \label{fig:pgsm_feature}
\end{figure}

\begin{table}[!t]
\begin{center}
\setlength{\tabcolsep}{4.5pt}
\small
\caption{Ablation study on different downsampling methods.}
\resizebox{\columnwidth}{!}{%
\begin{tabular}{l|ccccc}
\toprule[1pt]
Method & IoU$\uparrow$  & ${\rm P_d}\uparrow$ & ${\rm F_a}\downarrow$ & Params & FLOPs \\ \midrule
 w/ Pooling & 66.93 & 91.56 & 20.14 & 4.07 & 24.46 \\
 w/ Stride Conv & 67.10 & 91.58 & 23.22 & 4.26 & 25.06 \\  
 w/ SPDConv \cite{sunkara2022no} & \underline{67.87} & \underline{92.25} & \underline{15.63} & 5.07 & 29.41 \\  
 \rowcolor{gray!20}
 w/ MSAD  & \textbf{69.32} & \textbf{92.59} & \textbf{12.56} & 4.10 & 24.58 \\  
\bottomrule[1pt]
\end{tabular}
}
\label{tab:ab_down}
\end{center}
\end{table}

\begin{table*}[!t]
\centering
\scriptsize
\setlength{\tabcolsep}{6pt}
\caption{Scale-stratified IoU(\%) comparison on IRSTD-1k. SB-IoU denotes the average IoU over scale bins, and Avg. Rank denotes the average method rank across bins.}
\label{tab:scale_irstd1k}
\resizebox{\textwidth}{!}{%
\begin{tabular}{l|cccccc|cc}
\toprule[1pt]
Method & $[1,15]$ & $[16,31]$ & $[32,63]$ & $[64,127]$ & $[128,255]$ & $[256,\infty)$ & SB-IoU$\uparrow$ & Avg. Rank$\downarrow$ \\
\midrule
DNANet \cite{dnanet}
& 40.18 & 48.52 & 60.24 & 67.62 & 74.25 & \textbf{85.28} & 62.68 & 4.17 \\

MSHNet \cite{MSHNet}
& 43.63 & 56.81 & 61.43 & 65.39 & 77.96 & 83.57 & 64.80 & 3.67 \\

RPCANet \cite{wu2024rpcanet}
& 50.00 & 57.40 & 60.61 & \underline{73.82} & 75.61 & 51.62 & 61.51 & 3.50 \\

MSDANet \cite{zhao2025multi}
& \underline{50.45} & \textbf{68.71} & \textbf{66.56} & 71.04 & \textbf{81.86} & 77.26 & \underline{69.31} & \underline{2.00} \\

\rowcolor{gray!20}
RDANet (Ours)
& \textbf{56.05} & \underline{66.76} & \underline{66.08} & \textbf{75.78} & \underline{80.04} & \underline{84.91} & \textbf{71.60} & \textbf{1.67} \\
\bottomrule[1pt]
\end{tabular}
}
\end{table*}

\begin{table}[!t]
\centering
\small
\setlength{\tabcolsep}{4pt}
\caption{Background-stratified IoU(\%) comparison on IRSTD-1k. BG-Gap is the gap between the best and worst background-complexity groups.}
\label{tab:bg_irstd1k}
\resizebox{\columnwidth}{!}{%
\begin{tabular}{l|ccc|cc}
\toprule[1pt]
Method & Low & Medium & High & Mean$\uparrow$ & BG-Gap$\downarrow$ \\
\midrule
DNANet \cite{dnanet}
& 66.00 & 66.96 & 62.20 & 65.71 & 4.76 \\

MSHNet \cite{MSHNet}
& 70.87 & 69.79 & 64.44 & 67.87 & 6.43 \\

RPCANet \cite{wu2024rpcanet}
& 50.13 & 70.87 & 67.14 & 63.50 & 20.74 \\

MSDANet \cite{zhao2025multi}
& 69.19 & \textbf{75.78} & \underline{70.84} & \underline{71.89} & 6.59 \\

\rowcolor{gray!20}
RDANet (Ours)
& \textbf{75.92} & \underline{74.27} & \textbf{71.96} & \textbf{73.82} & \textbf{3.97} \\
\bottomrule[1pt]
\end{tabular}
}
\end{table}

\subsection{Discussion}
\label{sec:ablation_study}
We perform ablation studies on the IRSTD-1k dataset to evaluate the effectiveness of each component in RDANet, including MSAD, PGSM, and their design variants.

\paragraph{Overall Effectiveness of MSAD and PGSM}
We first evaluate the overall contribution of MSAD and PGSM in \cref{tab:ablation_msad_pgsm}. 
As shown in the \emph{Overall} block, the full RDANet achieves the best performance, reaching 73.82\% IoU, 93.60\% ${\rm P_d}$, and 7.67 ${\rm F_a}$, while introducing only marginal increases in parameters and FLOPs over the baseline. 

Removing either MSAD or PGSM consistently degrades performance, indicating that the two modules contribute complementary benefits: MSAD mainly improves structure preservation during downsampling, whereas PGSM enhances skip features for more stable target--background separation under scene variation. 
When both modules are removed, the performance drops to the baseline level, further confirming that the gain comes from their joint effect rather than incidental architectural changes. 
The fine-grained ablations further demonstrate that the internal components of both modules make distinct contributions. PixelUnshuffle and the SE gate each improve MSAD, whereas shift estimation and memory dropout enhance PGSM retrieval and false-alarm suppression. These components provide complementary improvements with negligible parameter and computational overhead.

We also compare several skip-refinement alternatives in \cref{tab:pgsm_skip_refinement}. SE improves IoU but increases false alarms, while CBAM and spatial prototype retrieval degrade both IoU and ${\rm F_a}$. Frequency retrieval without alignment reaches 72.60\% IoU, and the complete PGSM further improves IoU to 73.82\% while reducing ${\rm F_a}$ to 7.67, supporting the joint contribution of frequency-aware retrieval and phase-correlation alignment.

\begin{table}[!t]
\centering
\scriptsize
\setlength{\tabcolsep}{3pt}
\caption{Comparison of PGSM with alternative skip-refinement mechanisms on IRSTD-1k. Params and FLOPs are measured in M and G, respectively.}
\label{tab:pgsm_skip_refinement}
\resizebox{\columnwidth}{!}{%
\begin{tabular}{l|ccccc}
\toprule[1pt]
Variant & IoU$\uparrow$ & ${\rm P_d}\uparrow$ & ${\rm F_a}\downarrow$ & Params & FLOPs \\
\midrule
Direct skip & 71.32 & 92.86 & 11.26 & 4.09 & 25.12 \\
SE-enhanced skip & 72.31 & 93.27 & 16.97 & 4.10 & 25.13 \\
CBAM-enhanced skip & 70.92 & 92.93 & 30.38 & 4.10 & 25.16 \\
Prototype memory & 70.64 & 93.27 & 24.50 & 4.16 & 25.24 \\
Frequency memory w/o shift & 72.60 & 92.93 & 11.86 & 4.17 & 25.24 \\
\rowcolor{gray!20}
PGSM (full) & \textbf{73.82} & \textbf{93.60} & \textbf{7.67} & 4.17 & 25.24 \\
\bottomrule[1pt]
\end{tabular}
}
\end{table}

\paragraph{Mechanism Analysis of MSAD and PGSM}
We further investigate why the proposed modules improve detection performance. 
For MSAD, the frequency-spectrum visualization in \cref{fig:msad_vis} shows that anti-alias filtering suppresses peripheral high-frequency components while preserving the dominant low-frequency structure, indicating reduced aliasing and background leakage during resolution transformation. 
This observation is consistent with the downsampling comparison in \cref{tab:ab_down}, where MSAD outperforms conventional pooling and strided convolution with comparable computational overhead. 
The encoder feature visualization in \cref{fig:down_feature} further shows that MSAD maintains clearer and more localized target responses across stages, suggesting improved structural consistency after downsampling.
For PGSM, the feature maps in \cref{fig:pgsm_feature} show that memory-guided refinement makes skip features more target-focused and structurally complete, with clearer boundaries and fewer ambiguous background responses. 
Compared with directly propagating shallow encoder features, PGSM provides the decoder with more stable local contrast cues under complex scenes. 
Overall, MSAD mainly stabilizes feature transformation across scales, while PGSM improves scene-robust target enhancement in the skip pathway.

\paragraph{Analysis of Design Choices}
We further analyze several key design choices in RDANet. 
As shown in \cref{tab:ab_down}, conventional pooling and strided convolution yield inferior results, suggesting that standard downsampling is prone to structural distortion and background leakage in IRSTD. 
Although SPDConv \cite{sunkara2022no} alleviates this issue to some extent, it introduces noticeably higher computational cost. 
By contrast, MSAD achieves the best performance with only a modest increase in complexity, supporting the use of anti-aliased and structure-preserving downsampling.
We next examine the kernel choices in MSAD. As shown in \cref{tab:ab_msad_kernels}, the single- and dual-kernel results are non-monotonic: $K{=}\{5\}$ favors detection probability, whereas $K{=}\{7\}$ yields fewer false alarms than the other single-kernel settings. The complete $K{=}\{3,5,7\}$ configuration achieves the highest IoU and lowest ${\rm F_a}$, indicating that the gain comes from complementary smoothing ranges rather than simply stronger smoothing or more branches.

\begin{table}[!t]
\centering
\scriptsize
\setlength{\tabcolsep}{6pt}
\caption{Ablation study on blur-kernel configurations and fusion strategies in MSAD on IRSTD-1k.}
\label{tab:ab_msad_kernels}
\resizebox{\columnwidth}{!}{%
\begin{tabular}{l|l|ccc}
\toprule[1pt]
Kernels used & Fusion type & IoU$\uparrow$ & ${\rm P_d}\uparrow$ & ${\rm F_a}\downarrow$ \\
\midrule
$K{=}\{3\}$       & --            & 72.82 & 92.93 & 12.85 \\
$K{=}\{5\}$       & --            & 72.10 & \textbf{94.61} & 16.30 \\
$K{=}\{7\}$       & --            & 72.82 & 92.59 & \underline{9.11} \\
$K{=}\{3,5\}$     & Adaptive      & \underline{73.28} & 92.59 & 16.34 \\
$K{=}\{3,7\}$     & Adaptive      & 72.12 & \underline{93.60} & 13.61 \\
$K{=}\{5,7\}$     & Adaptive      & 71.74 & 91.25 & 13.95 \\
$K{=}\{3,5,7\}$   & Fixed average & 72.47 & 91.58 & 11.10 \\
\rowcolor{gray!20}
$K{=}\{3,5,7\}$   & Adaptive      & \textbf{73.82} & \underline{93.60} & \textbf{7.67} \\
\bottomrule[1pt]
\end{tabular}
}
\end{table}

\begin{table*}[!t]
\centering
\small
\setlength{\tabcolsep}{12pt}
\caption{Cross-dataset generalization and efficiency under two transfer settings.}
\label{tab:cross_dataset}
\resizebox{\textwidth}{!}{%
\begin{tabular}{l|ccc|ccc|ccc}
\toprule[1pt]
\multirow{2}{*}{Method} 
& \multicolumn{3}{c|}{IRSTD-1k $\rightarrow$ NUAA-SIRST} 
& \multicolumn{3}{c|}{NUAA-SIRST $\rightarrow$ IRSTD-1k}
& \multicolumn{3}{c}{Efficiency} \\
\cline{2-10}
& IoU$\uparrow$ & ${\rm P_d}\uparrow$ & ${\rm F_a}\downarrow$
& IoU$\uparrow$ & ${\rm P_d}\uparrow$ & ${\rm F_a}\downarrow$
& Params & FLOPs & FPS \\
\midrule
DNANet \cite{dnanet}   & 56.79 & 93.58 & 101.49  & 47.70    & 85.71    & 125.94   & 4.70 & 57.13  & 6 \\
MSHNet \cite{MSHNet}   & \underline{66.65} & 92.66 & 22.17  & \underline{54.54}    & 85.71    & \underline{61.64}    & 4.06 & 24.43  & 20 \\
RPCANet \cite{wu2024rpcanet} & 40.88 & 75.23 & 100.60 & 30.78  & \underline{87.63}    & 315.73    & 0.68 & 178.28 & 23 \\
MSDANet  \cite{zhao2025multi}     & 63.90 & \underline{92.70} & \underline{17.97} & 52.70    & 87.54    & 101.29    & 4.79 & 84.90  & 16 \\
\rowcolor{gray!20}
RDANet (Ours)   & \textbf{71.83} & \textbf{98.17} & \textbf{17.43} 
                & \textbf{60.74}   & \textbf{90.91}    & \textbf{60.24}    & 4.17 & 25.24  & 15 \\
\bottomrule[1pt]
\end{tabular}
}
\end{table*}

\begin{table*}[!t]
\centering
\small
\setlength{\tabcolsep}{2.5pt}
\caption{Re-training results on the augmented IRSTD-1k dataset. The gray values in parentheses indicate the change relative to the original result in the same metric/bin.}
\label{tab:retrain_scale}
\resizebox{\textwidth}{!}{%
\begin{tabular}{l|cccccc|ccc}
\toprule[1pt]
\multirow{2}{*}{Method} &
\multicolumn{6}{c|}{Pixel interval} &
\multirow{2}{*}{IoU$\uparrow$} &
\multirow{2}{*}{${\rm P_d}\uparrow$} &
\multirow{2}{*}{${\rm F_a}\downarrow$} \\
\cline{2-7}
& [1,15] & [16,31] & [32,63] & [64,127] & [128,255] & [256,$\infty$) &  &  &  \\
\midrule
DNANet \cite{dnanet}
& 37.66 \chg{-2.52}
& 52.95 \chg{+4.43}
& 58.49 \chg{-1.75}
& 63.32 \chg{-4.30}
& 72.63 \chg{-1.62}
& 84.27 \chg{-1.01}
& 63.74 \chg{-1.97}
& 89.12 \chg{-2.72}
& 19.59 \chg{+1.98} \\

MSHNet \cite{MSHNet}
& 44.56 \chg{+0.93}
& 57.96 \chg{+1.15}
& 60.07 \chg{-1.36}
& 64.60 \chg{-0.79}
& 76.38 \chg{-1.58}
& \textbf{85.64} \chg{+2.07}
& 67.56 \chg{-0.31}
& \textbf{93.88} \chg{+1.02}
& 9.49 \chg{+0.61} \\

RPCANet \cite{wu2024rpcanet}
& 35.85 \chg{-14.15}
& 51.92 \chg{-5.48}
& 58.77 \chg{-1.84}
& \underline{72.18} \chg{-1.64}
& 73.51 \chg{-2.10}
& 52.15 \chg{+0.53}
& 60.37 \chg{-3.13}
& 88.66 \chg{+1.03}
& 20.80 \chg{+12.15} \\

MSDANet \cite{zhao2025multi}
& \underline{54.91} \chg{+4.46}
& \underline{64.06} \chg{-4.65}
& \textbf{65.20} \chg{-1.36}
& 69.90 \chg{-1.14}
& \underline{78.66} \chg{-3.20}
& 83.72 \chg{+6.46}
& \underline{71.44} \chg{-0.45}
& 88.89 \chg{-5.39}
& \underline{8.62} \chg{-2.77} \\

\rowcolor{gray!20}
RDANet (Ours)
& \textbf{56.73} \chg{+0.68}
& \textbf{64.60} \chg{-2.16}
& \underline{65.10} \chg{-0.98}
& \textbf{75.96} \chg{+0.18}
& \textbf{79.41} \chg{-0.63}
& \underline{85.11} \chg{+0.20}
& \textbf{73.12} \chg{-0.70}
& \underline{91.58} \chg{-2.02}
& \textbf{6.28} \chg{-1.39} \\
\bottomrule[1pt]
\end{tabular}
}
\end{table*}

\begin{table}[!t]
\centering
\small
\setlength{\tabcolsep}{5.5pt}
\caption{Ablation study on the number of PGSM blocks.}
\label{tab:ab_pgsm}
\resizebox{\columnwidth}{!}{%
\begin{tabular}{c|ccccc}
\toprule[1pt]
PGSM Blocks  & IoU$\uparrow$  & ${\rm P_d}\uparrow$ & ${\rm F_a}\downarrow$ & Params & FLOPs \\ \midrule
 0 & 66.93 & 91.56 & 20.14 & 4.07 & 24.46 \\  
 1 & 69.14 & \textbf{92.86} & 15.35 & 4.07 & 24.70 \\  
 2  & 70.22 & 91.58 & \textbf{10.24} & 4.08 & 24.84 \\  
 3  & \underline{70.96} & \underline{92.25} & 12.43 & 4.09 & 24.98 \\    
 \rowcolor{gray!20}
 4   & \textbf{71.32} & \textbf{92.86} & \underline{11.26} & 4.13 & 25.12 \\  
\bottomrule[1pt]
\end{tabular}
}
\end{table}

\begin{table}[!t]
\centering
\scriptsize
\setlength{\tabcolsep}{3pt}
\caption{Sensitivity analysis of PGSM on IRSTD-1k. The default configuration is highlighted in gray.}
\label{tab:pgsm_sensitivity}
\resizebox{\columnwidth}{!}{%
\begin{tabular}{l|l|ccc}
\toprule[1pt]
Setting & Variant & IoU$\uparrow$ & ${\rm P_d}\uparrow$ & ${\rm F_a}\downarrow$ \\
\midrule
Memory size & $[64,32,16,8]$ & 70.93 & 91.58 & 11.12 \\
Memory size & $[128,64,32,16]$ & 72.65 & 93.27 & 13.40 \\
\rowcolor{gray!20}
Memory size & $[256,128,64,32]$ (default) & \textbf{73.82} & \textbf{93.60} & \textbf{7.67} \\
Memory size & $[512,256,128,64]$ & 71.46 & 92.26 & 9.75 \\
\midrule
Top-$K$ retrieval & $[4,4,4,2]$ & 70.86 & 93.60 & 17.82 \\
\rowcolor{gray!20}
Top-$K$ retrieval & $[8,8,8,4]$ (default) & \textbf{73.82} & \textbf{93.60} & \textbf{7.67} \\
Top-$K$ retrieval & $[16,16,8,4]$ & 73.08 & 93.27 & 10.74 \\
Top-$K$ retrieval & $[32,16,8,4]$ & 72.60 & 93.60 & 9.05 \\
\midrule
Patch size & $[4,4,4,4]$ & 72.86 & \textbf{94.95} & 21.81 \\
\rowcolor{gray!20}
Patch size & $[8,8,8,4]$ (default) & \textbf{73.82} & 93.60 & \textbf{7.67} \\
Patch size & $[16,8,8,4]$ & 73.36 & 93.94 & 11.35 \\
\bottomrule[1pt]
\end{tabular}
}
\end{table}

\begin{figure}[ht]
    \centering
    \includegraphics[width=0.95\linewidth]{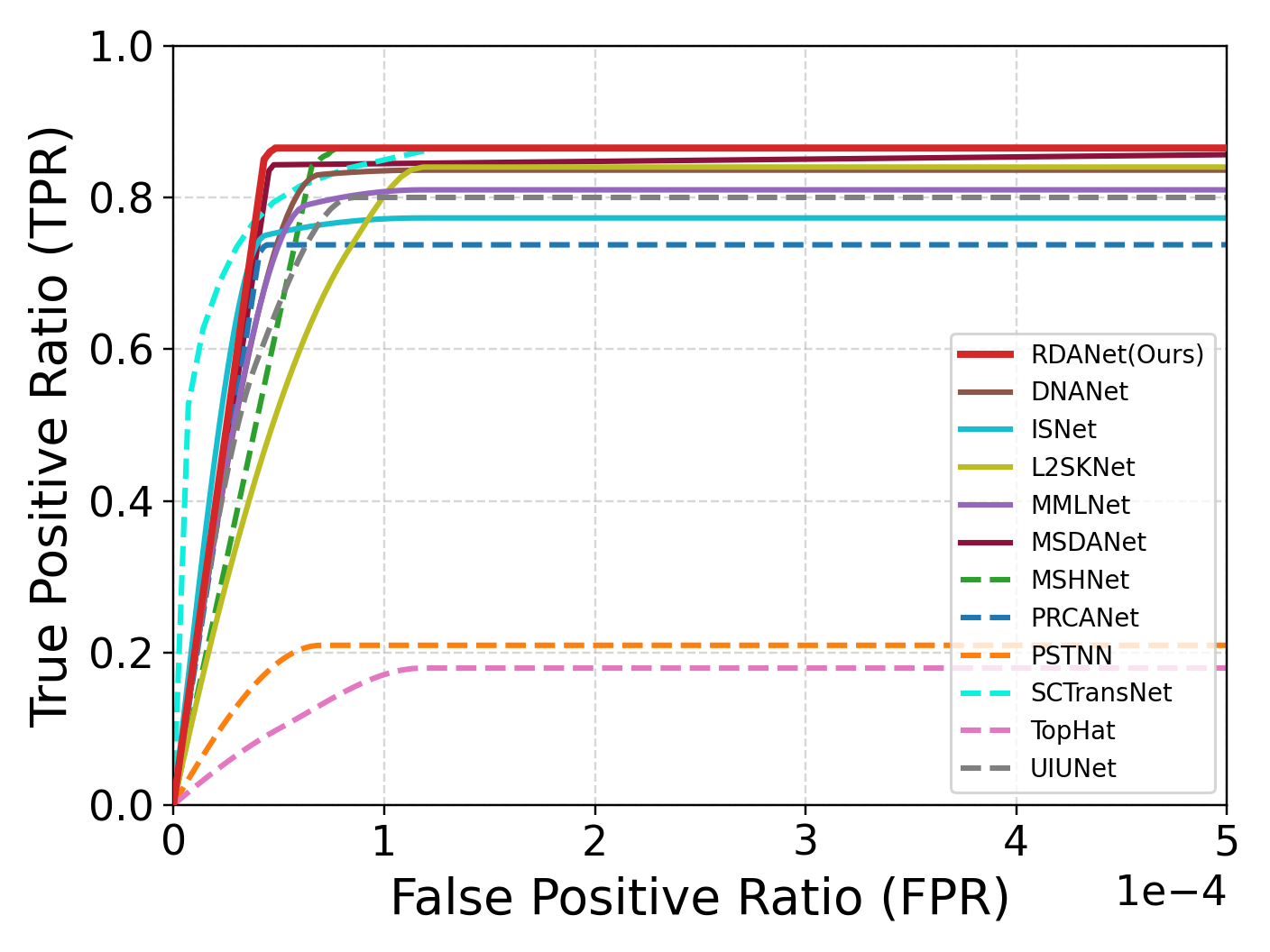}
    \caption{ROC curves on the IRSTD-1k dataset.}
    \label{fig:ROC}
\end{figure}

For PGSM, \cref{tab:ab_pgsm} shows that increasing the number of refinement blocks from one to four steadily improves IoU. \Cref{tab:pgsm_sensitivity} evaluates memory size, top-$K$ retrieval, and patch size; the default settings, $[256,128,64,32]$, $[8,8,8,4]$, and $[8,8,8,4]$, give the best overall balance, while alternative settings generally reduce IoU or increase ${\rm F_a}$.

\paragraph{Generalization and Robustness Analysis}
We further evaluate the robustness of RDANet in terms of cross-dataset transfer, scale-distribution changes, and the detection--false-alarm trade-off. First, the cross-dataset results in \cref{tab:cross_dataset} show that RDANet achieves the best performance in both transfer settings: testing IRSTD-1k-trained models on NUAA-SIRST and testing NUAA-SIRST-trained models on IRSTD-1k, both without fine-tuning. These results demonstrate stronger generalization to unseen scenes and background statistics. Moreover, RDANet improves both IoU and ${\rm P_d}$ while maintaining a low ${\rm F_a}$, indicating that its gains do not result from increased detection sensitivity at the cost of additional false alarms.

Second, we investigate whether the observed scale robustness can be attributed solely to rebalancing the training data. As shown in \cref{fig:teaser}, the original IRSTD-1k training set contains few samples in the largest target-size interval. To alleviate this imbalance, we construct 30 additional training images for the $[256,\infty)$ interval by compositing large-target masks from existing training samples onto background regions cropped from other images. This strategy preserves realistic target shapes while increasing scene diversity. We then retrain all methods under the same setting and report the scale-wise results in \cref{tab:retrain_scale}. Although several methods improve in certain large-target intervals, their performance continues to fluctuate across scales. In contrast, RDANet remains competitive across most intervals and achieves the best overall balance among IoU, detection probability, and false-alarm rate. This indicates that its scale robustness arises not only from data rebalancing but also from more stable feature representations across different target-size distributions.

Finally, the ROC curves in \cref{fig:ROC} further demonstrate the balance between detection sensitivity and false-alarm suppression. RDANet achieves a more favorable trade-off than the competing methods and maintains higher true-positive rates at low false-positive rates. Overall, these results demonstrate that RDANet provides strong benchmark performance together with improved generalization to unseen datasets, changes in target-scale distributions, and practical false-alarm constraints.

\paragraph{Limitations}
Despite its strong overall performance, RDANet still has several limitations. 
First, as shown in \cref{tab:scale_irstd1k,tab:scale_result}, it does not always achieve the best results in the smallest target interval.
This may be because the anti-alias filtering in MSAD slightly weakens extremely subtle target responses, while PGSM tends to favor more stable local structures. 
Second, although RDANet improves robustness under scene variation and cross-dataset transfer, its effectiveness still depends on the scale coverage and scene diversity of the training data. 
When extreme target sizes or background conditions are underrepresented, the learned prototypes may remain insufficiently representative. 
Third, as indicated in \cref{tab:cross_dataset}, RDANet maintains a favorable accuracy--efficiency trade-off, but it is not the fastest method, since the memory-guided skip refinement introduces additional computation compared with lighter baselines. 
Future work will explore finer-grained scale-adaptive enhancement, more discriminative prototype learning, and more efficient memory-guided designs.

\section{Conclusion}
In this paper, we revisit the problem of relative degradation in infrared small target detection, where detector performance becomes unstable as target scales expand or background conditions vary. 
We attribute this issue to insufficient structural preservation during resolution transformation and fragile local contrast modeling under scene variation. 
To address this issue, we propose RDANet, a simple encoder--decoder framework equipped with MSAD and PGSM.
MSAD reduces aliasing and background leakage while preserving target morphology, and PGSM refines skip features to provide more stable local contrast cues across different scenes. 
Together, these two components improve feature robustness across both target scales and background conditions.
Extensive experiments show that RDANet achieves competitive quantitative performance and more stable qualitative results across challenging scenes.
Further analyses verify its advantages in scale stability, scene robustness, and cross-dataset generalization. 
In future work, we will explore finer-grained scale-adaptive enhancement and more efficient memory interaction mechanisms for more challenging infrared scenes.

\bibliographystyle{IEEEtran}
\bibliography{references}

\begin{IEEEbiography}[{\includegraphics[width=1in,height=1.25in,clip,keepaspectratio]{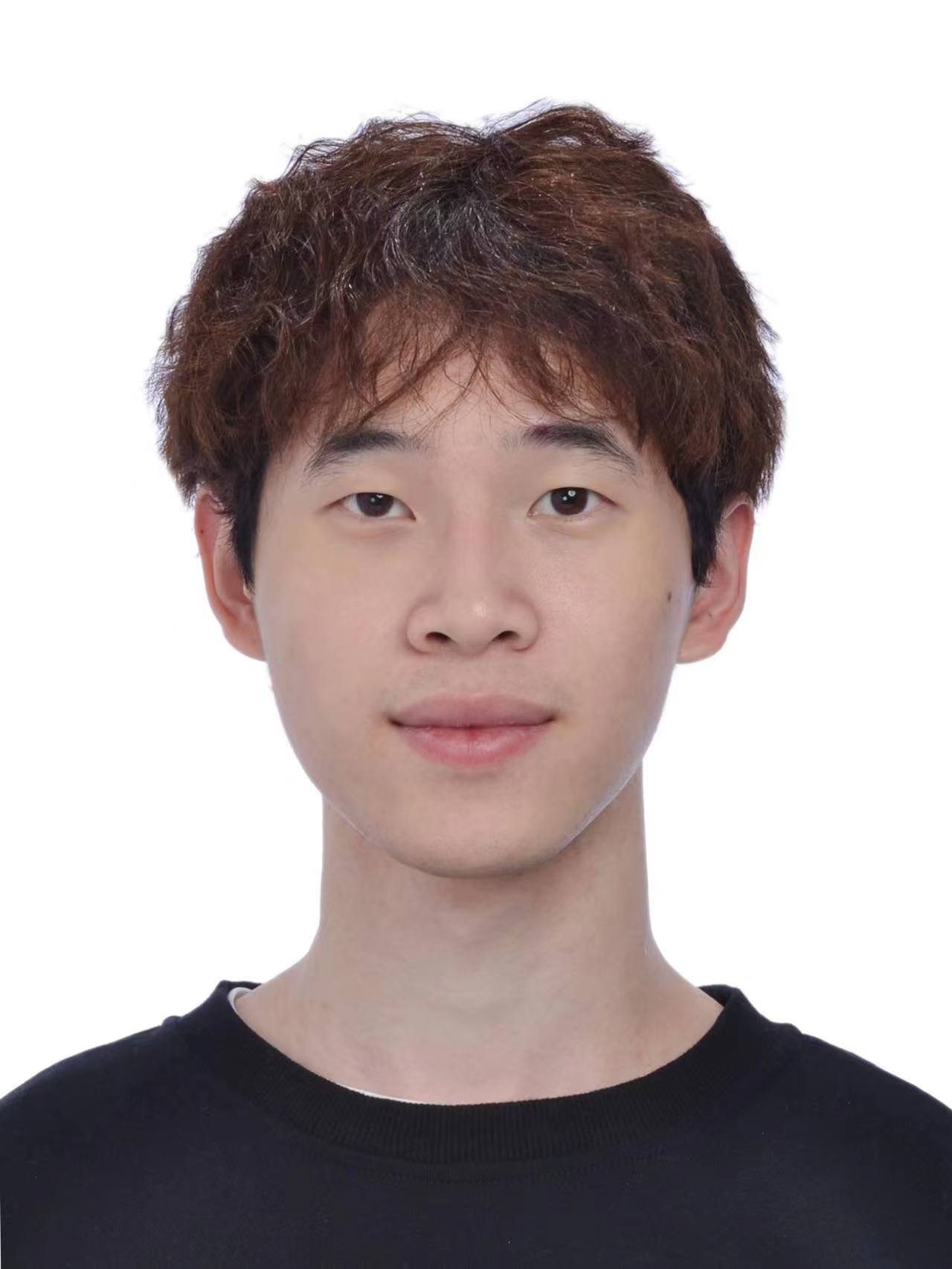}}]{Rui Liu}
	received the B.S. degree in Software Engineering in 2022 and the M.S. degree in Computer Technology in 2025, both from Beijing Institute of Technology, Beijing, China. He is currently pursuing the Ph.D. degree in Computer Science and Technology with the School of Computer Science and Technology, Beijing Institute of Technology. His research interests include object detection and remote sensing.
\end{IEEEbiography}
\begin{IEEEbiography}[{\includegraphics[width=1in,height=1.25in,clip,keepaspectratio]{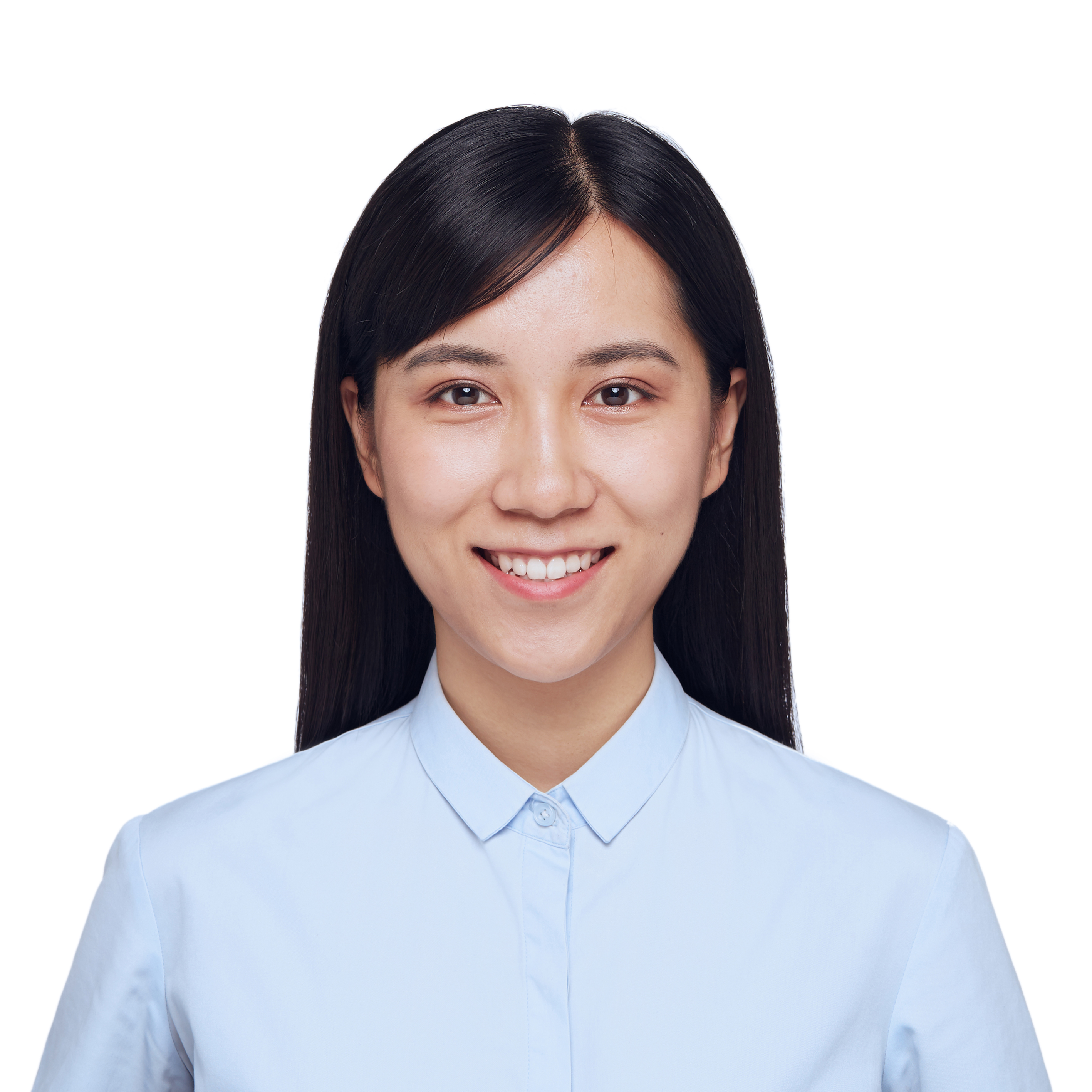}}]{Jing Nie}
	holds an M.S. degree and is currently an Assistant Researcher with the Beijing Institute of Remote Sensing Information, Beijing, China. Her research interests include intelligent satellite information processing and thermal infrared remote sensing data processing.
\end{IEEEbiography}
\begin{IEEEbiography}[{\includegraphics[width=1in,height=1.25in,clip,keepaspectratio]{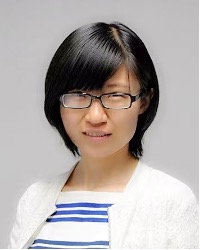}}]{Ying Fu}
	received the B.S. degree in Electronic Engineering from Xidian University in 2009, the M.S. degree in Automation from Tsinghua University in 2012, and the Ph.D. degree in Information Science and Technology from the University of Tokyo in 2015. 
	She is currently a professor at the School of Computer Science and Technology, Beijing Institute of Technology. Her research interests include physics-based vision, image and video processing, and computational photography. 
\end{IEEEbiography}

\end{document}